\documentclass[11pt]{article}
\usepackage{nodalida2021}
\usepackage{times}
\usepackage{hyperref}
\usepackage{url}
\usepackage{latexsym}
\usepackage{float}

\aclfinalcopy 
\def\aclpaperid{2} 

\title{Automatic Classification of Human Translation and Machine Translation: A Study from the Perspective of Lexical Diversity}

\author{Yingxue Fu \\
  School of Computer Science  \\
  University of St Andrews\\
  KY16 9SX, UK \\
  {\tt yf30@st-andrews.ac.uk} \\\And
  Mark-Jan Nederhof \\
  School of Computer Science  \\
  University of St Andrews\\
  KY16 9SX, UK \\
  }

\date{}
\begin{document}
\maketitle
\begin{abstract}
By using a trigram model and fine-tuning a pretrained BERT model for sequence classification, we show that machine translation and human translation can be classified with an accuracy above chance level, which suggests that machine translation and human translation are different in a systematic way. The classification accuracy of machine translation is much higher than of human translation. We show that this may be explained by the difference in lexical diversity between machine translation and human translation. If machine translation has independent patterns from human translation, automatic metrics which measure the deviation of machine translation from human translation may conflate difference with quality. Our experiment with two different types of automatic metrics shows correlation with the result of the classification task. Therefore, we suggest the difference in lexical diversity between machine translation and human translation be given more attention in machine translation evaluation.    

\end{abstract}

\section{Introduction}
The initial interest in and support for machine translation (MT) stem from visions of high-speed and high-quality translation of arbitrary texts~\citep{slocum1985survey}, but machine translation proves to be more difficult than initially imagined. In recent years, progress has been made in MT research and development, and it is claimed that MT achieves human parity in some tasks~\citep{wu2016google,hassan2018achieving,popel2020transforming}. However, these statements are challenged by other researchers and remain open to debate~\citep{laubli2018has, toral2018attaining,toral2020reassessing}.

The typical automatic approach to evaluating MT is to compare a machine translated text with a reference translation. The assumption is that the closer a machine translation is to a professional human translation, the better it is~\citep{papineni2002bleu}. Automatic metrics for MT are developed based on this assumption. Human translation (HT) is treated as gold standard and the deviation from it is transformed into a measure of translation quality of MT.

Many studies have shown that translated texts are different from originally written texts~\citep{baroni2006new, ilisei2010identification}. The typical method used for the identification of translationese is automatic classification of translated texts and originally written texts~\citep{baroni2006new}. There are some studies that compare translation varieties such as professional and student translations and post-edited MT~\citep{kunilovskaya2019translationese,toral2019post,popovic2020differences}. While surface linguistic features and simple machine learning techniques are capable of classifying translated texts and originally written texts with high accuracy, it is difficult to use the same method to classify translation varieties, with the accuracy being barely over the chance level~\citep{kunilovskaya2019translationese,rubino2016information}. 

When comparing translation varieties, MT is used as a translation variety independent of HT or other translation varieties in some studies~\citep{toral2019post}. Different from the conventional practice of MT evaluation that treats HT as the gold standard, some studies adopt a descriptive approach to comparing MT and HT~\citep{bizzoni-etal-2020-human,ahrenberg2017comparing,vanmassenhove-etal-2019-lost}. Among these studies,~\citet{bizzoni-etal-2020-human} find that MT shows independent patterns of translationese and it resembles HT only partly. This implies that MT may be different from HT in a systematic way, and 
it remains a question as to whether the deviation of MT from HT is a reliable measure of the quality of MT, and whether the current automatic metrics conflate differences between HT and MT with the quality of MT.

According to research by~\citet{toral2019post}, translation varieties differ in multiple ways. Based on research by~\citet{vanmassenhove-etal-2019-lost}, we focus on lexical diversity in our experiments. 

We try to answer three questions in this study: 
\begin{itemize}
    \item Can MT and HT be classified automatically with an accuracy above the chance level? 
    \item In what way does lexical diversity influence the classification result?
    \item Are the results of automatic metrics influenced by the difference in lexical diversity between HT and MT? 
\end{itemize}

\section{Related Work}
As our study essentially involves comparing translation varieties, we present an overview of 
previous studies that compare originally written texts and translations, other translation varieties, and HT and MT. 
\subsection{Comparing Originally Written Texts and Translations}
Translated texts show distinctive features which make them different from originally written texts. These features are typically studied under the framework of translationese.~\citet{gellerstam1986translationese} is the first to use this term to refer to the "fingerprints" that the source text leaves on the translated text. This notion is developed by Baker, who proposes the idea of universals of translation. As suggested by~\citet{baker1993corpus}, universals of translation are linguistic features that typically occur in translated texts 
as opposed to originally written texts, and these features are independent of the specific language pairs. Automatic means to distinguish translated texts and originally written texts have been developed and generally achieve high accuracy~\citep{baroni2006new, ilisei2010identification,lembersky2012language, rabinovich2015unsupervised}. Meanwhile, computational approaches~\citep{teich2003cross, volansky2015features} contribute evidence for some translation universals. 

\subsection{Comparing Translation Varieties}
Compared with the considerable amount of research on identifying translationese, the differences between translation varieties are less studied.

~\citet{rubino2016information} perform the classification between originally written texts and translations as well as between professional and student translations. They use surface features and distortion features which are inspired by quality estimation tasks, and surprisal and complexity features which are derived from information theory. Their experiment shows that originally written texts and professional translations are different mainly in terms of sequences of words, part-of-speech and syntactic tags, and originally written texts are closer to professional translations than to student translations. While the originally written texts and translations can be classified with high accuracy, automatic classification of different translation varieties is a more challenging task. Professional translations and student translations can only be classified with an accuracy barely above 50\%. 

This finding is consistent with the result of a study by~\citet{kunilovskaya2019translationese}. While morpho-syntactic features can be used to distinguish translations from non-translations with high accuracy, the performance of the same algorithm on classifying professional and student translations only slightly exceeds the chance level.

The differences of translations authored by human translators with different expertise and native languages are studied by~\citet{popovic2020differences}. Similar to other studies on distinguishing originally written texts from translated texts or comparing translation varieties, surface text features at word and part-of-speech levels are used. It concludes by suggesting that detailed information about the reference translation including translator information be provided in the scenario of MT evaluation. 

~\citet{toral2019post} compares post-edited MT with HT in terms of lexical variety, lexical density, sentence length ratio and part-of-speech sequences. The research shows that post-edited MT has lower lexical diversity and lower lexical density than HT, which is linked to the translation universal of simplification, and post-edited MT is more normalized and has greater 
interference from the source text (in terms of sentence length and part-of-speech sequences) than HT.

\subsection{Comparing MT and HT}

While the number of studies on comparing translation varieties is much smaller than on the identification of translationese, there are even fewer studies that explore the differences between MT and HT.

~\citet{ahrenberg2017comparing} compares MT and HT by means of automatically extracted features and statistics obtained through manual examination. By comparing the shifts (i.e. deviation from literal translation) and word order changes, he finds that HT contains twice as many word order changes. Meanwhile, 
an analysis of the number and types of edits required to give the machine translated text publishable quality is made. He argues that MT is likely to retain interference from the source text even after post-editing, and the machine translated text is more similar to the source text than the human translated text in many ways, including sentence length, information flow and structure. 

Research by~\citet{vanmassenhove-etal-2019-lost} shows another aspect where MT differs from HT. Three MT systems based on different architectures are trained. The lexical diversity of the translations of the MT systems is measured with three metrics including type/token ratio, Yule’s K, and measure of textual lexical diversity (MTLD). It is found that the output of neural machine translation (NMT) systems has a loss of lexical diversity compared with the human translated text. The reason for this phenomenon is that the advantage of NMT systems over statistical machine translation (SMT) systems in terms of learning over the entire sequence is obtained at the expense of discarding less frequently occurring words or morphological forms. This finding is consistent with the research 
by~\citet{toral2019post}, who observes that the lexical variety of post-edited MT is lower than of HT and the lexical variety of MT is lower than of post-edited MT, which is attributed to the tendency of MT to choose words used more frequently in the 
training data~\citep{farrell2018machine}. 

~\citet{bizzoni-etal-2020-human} study the differences between HT and MT in relation to the original texts. Part-of-speech perplexity and a syntactic distance metric are used to measure the differences between translations in written and spoken forms and produced by different types of MT systems. It is found that MT shows structural translationese, but the translationese of MT follows independent patterns that need further understanding. 

\section{Experiment}
We adopt two approaches for classifying MT and HT: developing a trigram language model with Witten-Bell smoothing and fine-tuning a pretrained BERT model for sequence classification from the Transformers library~\citep{wolf-etal-2020-transformers}. 

\subsection{Data}
The dataset is from the News commentary parallel corpus v13~\citep{tiedemann2012parallel} provided in the WMT2018 shared task\footnote{\url{http://www.statmt.org/wmt18/translation-task.html}}. We use Google Translate\footnote{\url{https://translate.google.co.uk}} to obtain the corresponding machine translation. 

The language pairs used in the experiment, the number of sentences for each language pair and the average sentence length for HT and MT are presented in Table~\ref{statistics}.

\begin{table}[H]
\centering
\begin{tabular}{|l|p{1.8cm}|p{1.4cm}|p{1.4cm}|}
\hline  
&\textbf{Number of sentences} 
&\textbf{MT avg sentence length} 
&\textbf{HT avg sentence length} \\ 
\hline
CS-EN & 30384 & 26.33 & 25.83\\ \hline
DE-EN & 30345 & 26.61 & 26.15 \\ \hline
RU-EN & 30387 & 28.00 & 27.51 \\ 
\hline
\end{tabular}

\caption{\label{statistics}Statistics of the dataset: translations from Czech, German and Russian to English.}
\end{table}

\subsection{Classifying HT and MT}
\subsubsection{Trigram Model}
We train two trigram models on the HT and MT training sets. Let $p_{MT}$ denote the trigram model trained on MT sentences, and $p_{HT}$ the model trained on HT sentences. A sentence $s$ is classified as MT if $p_{MT}(s)>p_{HT}(s)$ and as HT otherwise. If $s$ is from the HT test set and classified as HT, we count it as a success, and the same goes for the case when $s$ is from the MT test set and classified as MT. The classification accuracy is obtained by dividing the number of correct classifications by the total number of sentences in the respective test set. Since the two classes are balanced, accuracy is an appropriate metric. 

The result is shown in Table~\ref{n-gramorig}.

\begin{table}[H]
\begin{center}
\begin{tabular}{ |p{1.5cm} |p{1.5cm}| p{1.5cm} | }
\hline
\multicolumn{3}{|c|}{\textbf{CS-EN}} \\ \hline
              Total  &\ MT  &\ HT  \\ \hline
              0.69 &\ 0.79  &\ 0.58   \\ \hline

\multicolumn{3}{|c|}{\textbf{DE-EN}} \\ \hline
              Total  &\ MT  &\ HT  \\ \hline
              0.66 &\ 0.75  &\ 0.57   \\ \hline

\multicolumn{3}{|c|}{\textbf{RU-EN}} \\ \hline
             Total  &\ MT  &\ HT  \\ \hline
             0.67 &\ 0.76  &\ 0.58   \\ \hline
\end{tabular}
\end{center}
\caption{\label{n-gramorig}Classification accuracy of the trigram model.}
\end{table}

From Table~\ref{n-gramorig} it is clear that HT and MT can be classified automatically with an accuracy above the chance level. However, it is noticeable that MT can be classified with higher accuracy than HT.  

Based on research by~\citet{vanmassenhove-etal-2019-lost} and~\citet{toral2019post}, this imbalance in classification accuracy may be partly explained by the higher lexical diversity of HT, so that $p_{HT}$ is a probability distribution over sentences composed of a larger set of words than in the case of $p_{MT}$, thereby typically assigning a lower probability to any particular sentence, regardless of whether it is from MT or from HT. 

From Table~\ref{statistics}, it can be seen that the difference in average sentence length between MT and HT is only around 0.5. Therefore, we assume that the influence of sentence length is not significant in this study.  

\subsubsection{BERT Model}

We apply the BERT model on the same dataset, which is divided into training, test and validation sets by the ratio of 70\%, 10\% and 20\%. The sentences are padded to the maximum length of sentences in the dataset. We find that the pretrained BERT model for sequence classification achieves higher accuracy and lower loss in the first epoch. The result is shown in Table~\ref{bert-confmatrix}. 

\begin{table}[t]
\begin{center}
\begin{tabular}{|p{1.5cm} |p{1.5cm}| p{1.5cm}|}
\hline
\multicolumn{3}{|c|}{\textbf{CS-EN}} \\ \hline
              Total  &\ MT  &\ HT  \\ \hline
              0.78 &\ 0.90  &\ 0.66   \\ \hline

\multicolumn{3}{|c|}{\textbf{DE-EN}} \\ \hline
              Total  &\ MT  &\ HT  \\ \hline
              0.78 &\ 0.87  &\ 0.69   \\ \hline

\multicolumn{3}{|c|}{\textbf{RU-EN}} \\ \hline
             Total  &\ MT  &\ HT  \\ \hline
             0.78 &\ 0.90  &\ 0.65   \\ \hline
\end{tabular}
\end{center}
\caption{\label{bert-confmatrix}Classification accuracy of the BERT model.}
\end{table}

From Table~\ref{bert-confmatrix}, it can be seen that fine-tuning the pretrained BERT model for sequence classification can achieve higher accuracy for this task than the trigram model. 
Moreover, we can see the same pattern of imbalance in classification accuracy between MT and HT. Similar to the case of the trigram model, we hypothesize that it is because greater lexical diversity makes HT more difficult to classify correctly than MT. 

\subsection{Changing Lexical Diversity}

To investigate further whether differences in lexical diversity could be the reason for the observed imbalance in the classification accuracy of MT and HT, 
we manipulate the lexical diversity of the two.
As the lexical diversity of HT is generally higher than of MT~\citep{vanmassenhove-etal-2019-lost, toral2019post}, we reduce the lexical diversity of HT until it becomes close to or lower than MT, and for comparison, we also reduce the lexical diversity of MT.

\subsubsection{Method of Changing Lexical Diversity}

 Our general strategy of reducing lexical diversity is to replace rare words with words that are close to them in a vector space. First, we find rare words based on the frequency of lemmas in the corpus. Since there are many numerals and proper names and it is difficult to find meaningful candidates to replace them in the vector space, we set token.like\_num and token.is\_oov in spaCy processing\footnote{\url{https://spacy.io}} to false. Among the remaining lemmas, those lemmas whose frequency is lower than a threshold will be 
considered to be rare words. 
We found that setting the frequency threshold to two is effective in reducing the lexical diversity.

Second, we choose words whose vectors are close to the rare words from the pretrained GloVe embeddings~\citep{pennington-etal-2014-glove}, which are computationally less expensive than contextualized word embeddings like BERT. 
We found that the words which are closest to the rare words are not necessarily the optimal candidates in terms of part-of-speech or meaning, 
and so we choose the top three most similar words for each rare word. We convert the GloVe vectors into word2vec format with the gensim glove2word2vec API\footnote{\url{https://radimrehurek.com/gensim/scripts/glove2word2vec.html}} and set restrict\_vocab to 30000 in the most\_similar function\footnote{\url{https://radimrehurek.com/gensim/models/word2vec.html}} so that the search for the most similar words is limited to the top 30000 words in the pretrained embeddings. 
The vocabulary size 30000 was determined empirically. 

After this step, we apply a check on the fine-grained tags of the rare words and the fine-grained tags of the respective three candidates, the tags being obtained with spaCy \footnote{\url{https://spacy.io/api/token\#attributes}}
and containing more information than the coarse-grained part-of-speech tags from the Universal POS tag set\footnote{\url{https://universaldependencies.org/docs/u/pos/}}. The candidates with the same tags as the rare words will be chosen.
Where there is more than one matched candidate, only the first is chosen, and when there are no matched candidates after the check, the rare words will not be replaced. In this way, we obtain texts with modified lexical diversity. 
For ease of reference,
modified HT texts will be referred to as $HT\_modf$, modified MT texts will be referred to as $MT\_modf$, original HT texts as $HT\_orig$ and original MT texts as $MT\_orig$. 

To compute the lexical diversity of the texts, based on research by~\citet{mccarthy2010mtld} and~\citet{vanmassenhove-etal-2019-lost}, we choose the measure of textual lexical diversity (MTLD)~\citep{mccarthy2005assessment}, which is reasonably robust to text length difference. We refer those interested in the specific computation and statistical significance of MTLD to~\citet{mccarthy2010mtld}. The lexical diversity of the texts is presented in Table~\ref{mtldtable}.

\begin{table}[H]
\centering
\begin{tabular}{|p{2cm} |p{2cm}| p{2cm}|}
\hline  
MTLD & \textbf{Original} & \textbf{Modified} \\ 
\hline
CS\_MT & 62.02 & 43.00 \\
\hline
CS\_HT & 63.80 & 43.04 \\
\hline
DE\_MT & 62.53 & 42.44 \\
\hline
DE\_HT & 64.59 & 42.76 \\
\hline
RU\_MT & 61.06 & 42.66 \\
\hline
RU\_HT & 64.51 & 43.05 \\
\hline
\end{tabular}
\caption{\label{mtldtable}MTLD of the 
original texts and of the modified texts.}
\end{table}

From Table~\ref{mtldtable}, it can be seen that the MTLD values of HT texts are generally higher than of MT texts, which is consistent with the result of previous studies~\citep{vanmassenhove-etal-2019-lost, vanmassenhove-etal-2021-machine, toral2019post}. With our method, the difference in MTLD value between MT and HT texts is reduced. 

\subsubsection{Experimental Result of Trigram Model}
We conduct another set of binary classification experiments on the original and modified MT and HT texts paired in different ways.  For example,
``$MT\_modf$ \& $HT\_modf$" in the following tables means that the binary classification is performed on the modified MT text and the modified HT text. The result of the trigram model is shown in Table~\ref{newtrigram}. 
For comparison, the results from Table~\ref{n-gramorig} are repeated in the lines
$MT\_orig$ \& $HT\_orig$. 

\begin{table}[H]
\begin{center}
\begin{tabular}{|l|l|l|l|}
\hline
\multicolumn{4}{|c|}{\textbf{CS-EN}} \\ \hline
  Accuracy & Total  & MT  & HT  \\ \hline
${MT\_orig}$ \& ${HT\_orig}$  & 0.69 & 0.79  & 0.58 \\ \hline
${MT\_modf}$ \& ${HT\_modf}$  & 0.69  & 0.77 & 0.61 \\ \hline 
${MT\_orig}$ \& ${HT\_modf}$  & 0.69  & 0.56 & 0.83 \\ \hline 

\multicolumn{4}{|c|}{\textbf{DE-EN}} \\ \hline
  Accuracy & Total  & MT  & HT  \\ \hline
${MT\_orig}$ \& ${HT\_orig}$  & 0.66 & 0.75  & 0.57 \\ \hline
${MT\_modf}$ \& ${HT\_modf}$  & 0.67  & 0.74 & 0.60 \\ \hline 
${MT\_orig}$ \& ${HT\_modf}$  & 0.67  & 0.52 & 0.82 \\ \hline 

\multicolumn{4}{|c|}{\textbf{RU-EN}} \\ \hline
  Accuracy & Total  & MT  & HT  \\ \hline
${MT\_orig}$ \& ${HT\_orig}$  & 0.67 & 0.76  & 0.58 \\ \hline
${MT\_modf}$ \& ${HT\_modf}$  & 0.67  & 0.75 & 0.59 \\ \hline 
${MT\_orig}$ \& ${HT\_modf}$  & 0.67  & 0.52 & 0.82 \\ \hline 
\end{tabular}
\end{center}

\caption{\label{newtrigram}Binary classification of MT and HT by the trigram model under different combinations of MT and HT texts.}
\end{table}

From Table~\ref{newtrigram} in combination with Table~\ref{mtldtable}, we can see that when the difference in lexical diversity between MT and HT becomes smaller, the imbalance in classification accuracy is reduced, and the classification accuracy of MT goes down while the classification accuracy of HT goes up.

Since the lexical diversity of HT is generally higher than MT, we conduct an experiment where the lexical diversity of HT is significantly lower than MT, and the result is shown in the lines ${MT\_orig}$ \& ${HT\_modf}$. Under this condition, the classification accuracy of MT is much lower than HT. In this way, we reverse the previously observed trend that the classification accuracy of MT is higher than HT. Note that the overall classification accuracy does not change much in this experiment.  

\subsubsection{Experimental Result of BERT Model}

For fine-tuning the pretrained BERT model for sequence classification, similar experiments were done, with different combinations of MT and HT texts. Accuracies are presented in Table~\ref{newbertmodel}.

\begin{table}[H]
\begin{center}
\begin{tabular}{|l|l|l|l|}
\hline
\multicolumn{4}{|c|}{\textbf{CS-EN}} \\ \hline
  Accuracy & Total  & MT  & HT \\ \hline
\mbox{${MT\_orig}$ \& ${HT\_orig}$}  & 0.78 & 0.90  & 0.66 \\ \hline
${MT\_modf}$ \& ${HT\_modf}$  & 0.78  & 0.89 & 0.68 \\ \hline 
${MT\_orig}$ \& ${HT\_modf}$  & 0.82  & 0.91 & 0.73 \\ \hline 

\multicolumn{4}{|c|}{\textbf{DE-EN}} \\ \hline
  Accuracy & Total  & MT  & HT  \\ \hline
${MT\_orig}$ \& ${HT\_orig}$  & 0.78 & 0.87  & 0.69 \\ \hline
${MT\_modf}$ \& ${HT\_modf}$  & 0.78  & 0.86 & 0.71 \\ \hline 
${MT\_orig}$ \& ${HT\_modf}$  & 0.81  & 0.89 & 0.73 \\ \hline 

\multicolumn{4}{|c|}{\textbf{RU-EN}} \\ \hline
  Accuracy & Total  & MT  & HT  \\ \hline
${MT\_orig}$ \& ${HT\_orig}$  & 0.78 & 0.90  & 0.65 \\ \hline
${MT\_modf}$ \& ${HT\_modf}$  & 0.77  & 0.89 & 0.65 \\ \hline 
${MT\_orig}$ \& ${HT\_modf}$  & 0.81  & 0.95 & 0.68 \\ \hline 
\end{tabular}
\end{center}
\caption{\label{newbertmodel}Binary classification of MT and HT by the BERT model under different combinations of MT and HT texts.}
\end{table}
 
Similar to the trigram model, the classification accuracy of HT goes up 
in the case of CS-EN and DE-EN
and the classification accuracy of MT goes down a little, when the lexical diversity of MT and of HT are closer, as shown in 
the lines ${MT\_modf}$ \& ${HT\_modf}$, 
and when the lexical diversity of HT is much lower than MT, the classification accuracy of HT goes up, as shown in the 
lines ${MT\_orig}$ \& ${HT\_modf}$.
However, changing the difference in lexical diversity does not tend to decrease the classification accuracy of MT for the BERT model. Recall that with the trigram model, the classification accuracy of HT 
increases
while the classification accuracy of MT decreases. In contrast, with the BERT model, even when the lexical diversity of MT is much higher than HT, the overall classification accuracy and the separate classification accuracies of MT and HT all go up. The difference of the two models in terms of the classification accuracy of MT may be explained by the fact that the pretrained BERT model for sequence classification calculates cross-entropy loss for the classification 
task\footnote{\url{https://github.com/huggingface/transformers/blob/9aeacb58bab321bc21c24bbdf7a24efdccb1d426/src/transformers/modeling_bert.py}} while the trigram model results from relative frequency estimation. 
 
\subsection{Automatic Metrics}
We hypothesize that the performance of the two models in the binary classification task may be reflected in the result of MT metrics that are based on n-gram matching or that use contexualized embeddings.

Since BLEU is a commonly used metric based on n-gram matching, we test the performance of BLEU on the dataset to see if the difference in lexical diversity between MT and HT would influence the result. 
We calculate the corpus-level BLEU score for MT, as implemented in NLTK\footnote{\url{https://www.nltk.org/}}, using HT as reference.
The result is presented in Table~\ref{bleu}. 

\begin{table}[H]
\centering
\begin{tabular}{|l|p{1.5cm}| p{1.7cm}| p{1.6cm}|}
\hline  
 BLEU   & ${MT\_orig}$ \& ${HT\_orig}$  & ${MT\_modf}$ \& ${HT\_modf}$  & ${MT\_orig}$ \& ${HT\_modf}$ \\ 
\hline
\mbox{CS-EN} & 0.42 & 0.46 & 0.39 \\
\hline
\mbox{DE-EN} & 0.41 & 0.45 & 0.38 \\
\hline
\mbox{RU-EN} & 0.37 & 0.40 & 0.34 \\
\hline
\end{tabular}
\caption{\label{bleu}BLEU score.}
\end{table}

As can be seen from Table~\ref{bleu}, when the lexical diversity of MT is closest to HT, as shown by the column ${MT\_modf}$ \& ${HT\_modf}$, the MT BLEU score is the highest. When the lexical diversity of the reference is much lower than MT, as is the case in the column ${MT\_orig}$ \& ${HT\_modf}$, the MT BLEU score is the lowest. 
Much as in the discussion of the results of the trigram model, the difference in lexical diversity between MT and HT is a factor that needs to be taken into account when an n-gram matching based metric like BLEU is used for MT evaluation. 

The majority of automatic MT metrics developed in recent years such as BERTScore~\citep{zhang2019bertscore} and Yisi~\citep{lo2019yisi} adopt contextualized embeddings. Based on accessibility and performance, we choose MoverScore~\citep{zhao2019moverscore} 
as an example of a metric that uses BERT representations.
Since MoverScore is not a corpus-level metric, 
we calculate the average sentence-level score.
The result is presented in Table~\ref{moverscore}. 

\begin{table}[h]
\centering
\begin{tabular}{|p{1.1cm}|p{1.5cm}| p{1.7cm}| p{1.6cm}|}
\hline  
Mover-Score   & ${MT\_orig}$ \& ${HT\_orig}$  & ${MT\_modf}$ \& ${HT\_modf}$  & ${MT\_orig}$ \& ${HT\_modf}$ \\ 
\hline
\mbox{CS-EN} & 0.57 & 0.56 & 0.55 \\
\hline
\mbox{DE-EN} & 0.57 & 0.56 & 0.55 \\
\hline
\mbox{RU-EN} & 0.52 & 0.50 & 0.50 \\
\hline
\end{tabular}
\caption{\label{moverscore}MoverScore result for MT.}
\end{table}

The MoverScore result in Table~\ref{moverscore} shows a different pattern from the BLEU scores. The scores are basically inversely proportional to the overall accuracy of the binary classification task shown in Table~\ref{newbertmodel}. As the difference in MoverScore results under different combinations of MT and HT texts is small, more work is needed. 

\section{Conclusion and Future Work}

With the above experiments, we have shown that MT and HT can be classified with an accuracy above the chance level. The trigram model does not involve a machine learning algorithm but is capable of capturing the differences between MT and HT. By fine-tuning the pretrained BERT model for sequence classification, we obtain a higher accuracy for this task.

Similar to the identification of translationese, we may claim that MT and HT belong to different translation varieties. The result serves as supporting evidence for the study by~\citet{bizzoni-etal-2020-human}, which maintains that MT only resembles HT in part and often follows independent patterns. This finding calls into question the longstanding assumption in MT evaluation that the more similar an MT output is to a professional human translation, the better it is. If MT and HT are two translation varieties and have different patterns, it leaves room for doubt as to the legitimacy of evaluating MT by its similarity to HT.  

Moreover, there is a noticeable imbalance in the classification accuracy of HT and MT.
For the trigram model, while more than 70\% of the MT test sentences can be classified correctly, 
fewer than 60\% of the HT test sentences are classified correctly. This imbalance also exists in the experiment with the BERT model. 
Generally speaking, it is easier to correctly classify MT sentences than HT sentences. 

Based on previous studies and analysis from the probabilistic perspective, we consider lexical diversity as one of the major reasons for this imbalance in classification accuracy. We change the lexical diversity of the MT and HT texts and conduct another set of experiments with the same models. With the trigram model, if the difference in lexical diversity between MT and HT decreases, the imbalance in classification accuracy between the two is reduced, and we can reverse this imbalance in classification accuracy when the lexical diversity of MT is higher than HT. The result of the experiment with the BERT model shows a different pattern. An increase in classification accuracy of HT is accompanied by an increase in the classification accuracy of MT. This may be explained by the different ways of performing binary classification by the two models.    

The performance of automatic MT metrics based on n-gram matching, represented by BLEU in this study, and automatic metrics using BERT representations, such as MoverScore, is related to the result of the binary classification task with the two kinds of models. When the lexical diversity of HT is lower than MT, the MT BLEU score is the lowest and when the lexical diversity of HT is very close to MT, the MT BLEU score is the highest. The evaluation results given by MoverScore are basically inversely proportional to the classification accuracy of the BERT model. 
Therefore, we suggest the difference in lexical diversity between MT and the reference be given more attention in MT evaluation with automatic metrics.      

We are aware that there are other possible factors that may account for the phenomenon that HT is more likely to be classified as MT than the other way around.
In our experiment, we only manipulate one factor.
In future work, we intend to further study the independent patterns of MT compared with HT and investigate if the differences between MT and HT are related to the quality of MT. 
As differences in lexical diversity may influence automatic metrics for MT evaluation in
different ways, 
we plan to explore this phenomenon with other metrics, such as COMET~\citep{rei-etal-2020-comet}. 

\bibliographystyle{acl_natbib}
\bibliography{nodalida2021}

\end{document}